\documentclass{article}

\usepackage[nonatbib,preprint]{neurips_2024}

\usepackage[utf8]{inputenc} % allow utf-8 input
\usepackage[T1]{fontenc}    % use 8-bit T1 fonts
\usepackage{hyperref}       % hyperlinks
\usepackage{url}            % simple URL typesetting
\usepackage{booktabs}       % professional-quality tables
\usepackage{amsfonts,amssymb, amsmath}       % blackboard math symbols
\usepackage{nicefrac}       % compact symbols for 1/2, etc.
\usepackage{microtype}      % microtypography
\usepackage{xcolor}         % colors
\usepackage{graphicx}

\usepackage{colortbl}
\usepackage{subcaption}
\usepackage{enumitem}
\usepackage{multirow}
\usepackage{amsmath}
\usepackage{float}
\usepackage{pifont}       % \ding{xx}
\usepackage{bbding}       % \Checkmark,\XSolid,... (需要和pifont宏包共同使用)
\usepackage{fontawesome}  % \faCheck,\faTimes
\usepackage{wrapfig}
\usepackage{tcolorbox}
\usepackage{soul}
\usepackage{comment}
\usepackage{arydshln}
\usepackage{tikz}

\newcolumntype{P}[1]{>{\centering\arraybackslash}p{#1}}

\definecolor{codeblue}{rgb}{0.25,0.5,0.5}
\definecolor{myblue}{rgb}{0.88,0.98,1}
\definecolor{mygreen}{rgb}{0.92, 1.0, 0.92}
\definecolor{myred}{rgb}{1, 0.9, 0.9}
\definecolor{mygray}{gray}{0.93}
\definecolor{mygray1}{gray}{0.95}
\newcommand{\colorrect}[1]{\textcolor{#1}{\ding{110}}}
\definecolor{Highlight}{HTML}{E8F8F5}
\definecolor{midgreen}{HTML}{69c5a3}
\definecolor{midblue}{HTML}{6ca6cd}
\definecolor{darkgreen}{HTML}{77A351}
\definecolor{darkblue}{HTML}{74B8EF}

\definecolor{mygreen}{RGB}{93,173,85}

\makeatletter
\newcommand{\thickhline}{
    \noalign {\ifnum 0=`}\fi \hrule height 0.5pt
    \futurelet \reserved@a \@xhline
}

\newcommand*\circled[1]{\tikz[baseline=(char.base)]{
            \node[shape=circle,draw,inner sep=0.01pt] (char) {#1};}}

\title{TokenPacker: \\ Efficient Visual Projector for Multimodal LLM}

\author{Wentong Li$^{1*}$, Yuqian Yuan$^{1}$\thanks{Equal contribution.}~~,  Jian Liu$^{2}$, Dongqi Tang$^{2}$, Song Wang$^{1}$, \\
\textbf{Jie Qin$^{3}$,}
\textbf{~Jianke Zhu$^{1}$\thanks{Corresponding author.}~, ~ Lei Zhang$^{4}$} \\
  $^{1}$Zhejiang University \quad
  $^{2}$Ant Group  \quad 
  $^{3}$NUAA  \quad 
  $^{4}$The Hong Kong Polytechnical University
  }

\begin{document}

\maketitle

\begin{abstract}
The visual projector serves as an essential bridge between the visual encoder and the Large Language Model (LLM) in a Multimodal LLM (MLLM).  Typically, MLLMs adopt a simple MLP to preserve all visual contexts via one-to-one transformation. However, the visual tokens are redundant and can be considerably increased when dealing with high-resolution images, impairing the efficiency of MLLMs significantly. Some recent works have introduced resampler or abstractor to reduce the number of resulting visual tokens. Unfortunately, they fail to capture finer details and undermine the visual reasoning capabilities of MLLMs. In this work, we propose a novel visual projector, which adopts a coarse-to-fine scheme to inject the enriched characteristics to generate the condensed visual tokens. In specific, we first interpolate the visual features as a low-resolution point query, providing the overall visual representation as the foundation. 
Then, we introduce a region-to-point injection module that utilizes high-resolution, multi-level region-based cues as fine-grained reference keys and values, allowing them to be fully absorbed within the corresponding local context region. 
This step effectively updates the coarse point query, transforming it into an enriched one for the subsequent LLM reasoning.  Extensive experiments demonstrate that our approach compresses the visual tokens by 75\%$\sim$89\%, while achieves comparable or even better performance across diverse benchmarks with significantly higher efficiency.  The source codes can be found at \url{https://github.com/CircleRadon/TokenPacker}.

\end{abstract}

\section{Introduction}
With the rapid evolution in Large Language Models (LLM)~\cite{zheng2023vicuna,2023internlm,qwen,touvron2023llama,touvron2023llama2,openai2023gpt4,mixtral},  Multimodal Large Language Models (MLLMs)~\cite{liu2023llava,llava1.5, llavanext,bai2023qwen-vl,zhu2023minigpt4,chen2024howfar,dong2024internlm,reid2024gemini1_5,zhang2024mm} has witnessed a significant surge in vision-language understanding, reasoning, and interaction capabilities. This is achieved by projecting embeddings from a visual encoder into LLM to enable their visual perception of the world, where visual projector plays a crucial role to bridge the vision and language model.

In the framework of MLLMs, the LLM predominantly drives the entire computation cost, particularly since the visual encoder tends to be substantially smaller compared to the LLM. For instance, the widely used CLIP-ViT-Large~\cite{CLIP}, which features 0.3 billion parameters, stands in stark contrast to LLMs such as LLaMA~\cite{touvron2023llama} or Vicuna~\cite{Vicuna} with 7/8 billion or 13 billion parameters. Consequently, the efficiency of MLLMs is significantly affected by the number of resulting visual tokens from visual projector. Besides, the visual projector connects the vision and language models by translating visual features into visual tokens in a text embedding space that language model can interpret. Therefore, the quality of these visual tokens directly affects the overall efficacy of MLLM. In this work, we aim to investigate an effective visual projector for an MLLM that bridges the vision encoder and LLM with high quality, while making use of the fewer number of tokens possible.
\begin{figure}[t]
\begin{center}
\includegraphics[width=1.0\linewidth]{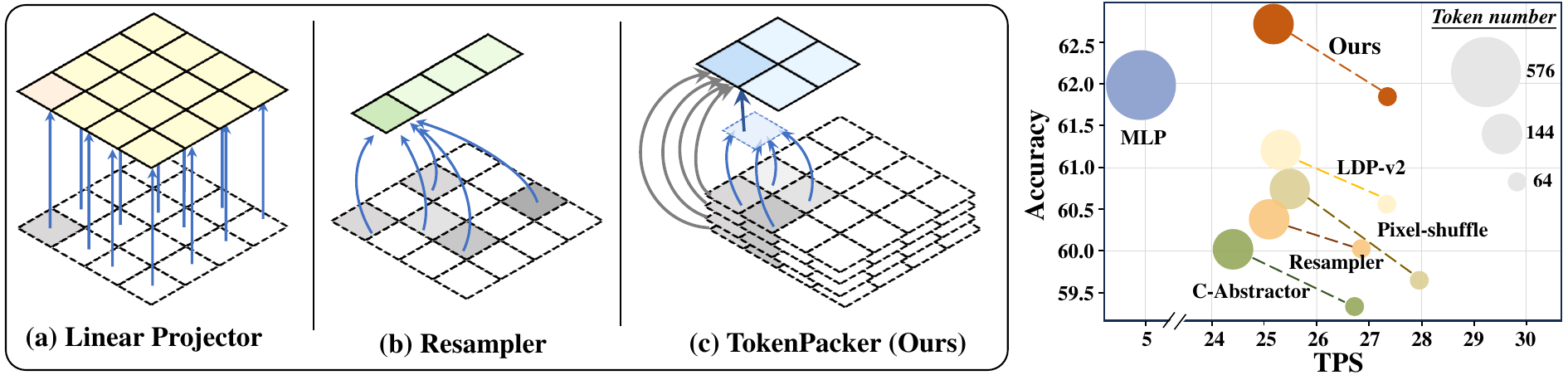} \end{center}
 \vspace{-5pt}
    \caption{(Left) Visual comparisons on typical projectors, including linear MLP~\cite{llava1.5} and Resampler~\cite{bai2023qwen-vl}. Our approach mines multi-level features in a local context region. (Right) Accuracy \textit{vs.} efficiency (TPS) comparisons with existing methods. Our TokenPacker shows a favorable performance against other counterparts. The accuracy is averaged across six benchmarks (see Table~\ref{tab:main_result_normal_resolution}).}
 \label{fig:projector_compare}
\vspace{-10pt}
 \end{figure}

Most of current works adopt either linear projector~\cite{liu2023llava, llava1.5} or resampler~\cite{bai2023qwen-vl, instructblip, ye2023mplug}. 
As for linear projector,  MLP projection~\cite{llava1.5} preserves all visual contexts via one-to-one transformation, which retains the detailed information having redundant tokens~\cite{bolya2022token,2024llava-prumerge}. More importantly, the number of visual tokens is significantly increased in dealing with high-resolution images or videos. 
As for another research line, resampler~\cite{bai2023qwen-vl} or Q-Former~\cite{BLIP2} leverage a group of learnable queries to explicitly controls the number of visual tokens and adopt the cross-attention layers to force extracting the most relevant visual cues from visual features. Some recent studies, \textit{e.g.} Abstractor~\cite{cha2023honeybee} or LDP ~\cite{mobilevlmv1, mobilevlmv2}, utilize convolution layers to encourage local interaction of visual features and generate the compressed tokens. 
Nonetheless, these methods inevitably lose the finer details information and sacrifice the visual reasoning capabilities of MLLMs. Besides, some methods directly transfer visual features from sequence dimension to channel dimension by a simple pixel shuffle~\cite{chen2024howfar} or nearby concatenation operation~\cite{dong2024internlm} to reduce the length of sequence. Although having preserved all information, it may destroy the structural characteristics of the visual feature itself.

In this work, we propose a novel visual projector, dubbed TokenPacker, which effectively packs the finer detailed information into compact visual token representations. 
Our TokenPacker aligns with a coarse-to-fine design, which injects enriched high-resolution characteristics into a coarse low-resolution one to generate the condensed visual tokens. Specifically, we initially interpolate  visual feature from the vision encoder as low-resolution point queries, which contain coarse and holistic characteristics of visual cues.  Then, we introduce a region-to-point injection module, which makes full use of high-resolution, multi-level CLIP features  to provide fine-grained candidate keys and values for reference. During this process, high-resolution  visual region details are encouraged to inject into the low-resolution point query to be updated within a local context region. This effectively enhances the coarse query and transforms it into a more enriched one for the subsequent LLM. As an extension, we further present an effective dynamic image slicing scheme to perform efficient high-resolution image understanding with our TokenPacker. 

Extensive experiments are conducted across diverse multimodal benchmarks to investigate the efficacy of our approach.  Notably, our TokenPacker can effectively reduce 75\% (576 \textit{vs.} 144)$\sim$89\% (576 \textit{vs.} 64) visual tokens in LLaVA-1.5~\cite{llava1.5} while achieving comparable or even better performance with significantly higher efficiency. As illustrated in Fig.~\ref{fig:projector_compare}, our method exhibits a more favorable superiority on accuracy and efficiency against other counterparts. Additionally, our approach consistently delivers competitive high-resolution   comprehension performance on a variety of multimodal tasks.

\section{Related Work}

\subsection{Multimodal Large Language Models (MLLMs)}

Large Language Models (LLMs)~\cite{touvron2023llama,touvron2023llama2,zheng2023vicuna,qwen,2023internlm,openai2023gpt4,mixtral,openai2023gpt4} have attracted considerable attention for their
remarkable capabilities across various linguistic tasks, such as question answering and text generation. 
This wave of interest has paved the way for the development of recent Multimodal Large Language Models (MLLMs)~\cite{zhang2024mm}, which integrate LLMs with visual encoders to enable an enriched comprehension and understanding of multimodal content. 
Innovative models like CLIP~\cite{CLIP} have significantly narrowed the gap between language processing and visual tasks, boosting the cross-modal applications. Early efforts such as Flamingo~\cite{flamingo} and BLIP-2~\cite{BLIP2}, have leveraged extensive datasets of image-text pairs to refine cross-modal alignment, substantially enhancing learning efficiency. 
This enhancement represents a notable advancement in the field of MLLMs, expanding the scope of applications by accommodating both text and imagery. 
In the recent year, a variety of MLLMs have gained prominence. Notable open-source examples include the LLaVA series~\cite{liu2023llava,llava1.5, llavanext}, MiniGPT-4~\cite{zhu2023minigpt4}, Qwen-VL~\cite{bai2023qwen-vl}, CogVLM~\cite{wang2023cogvlm}, Shikra~\cite{chen2023shikra}, InternLM-XComposer~\cite{dong2024ixc2} and  
among others~\cite{chen2023internvl,lu2024deepseek}.  The emergence of proprietary commercial MLLMs marks a pivotal shift in the landscape, as seen with OpenAI's GPT-4V~\cite{gpt4v} and Google’s Gemini series~\cite{team2023gemini, reid2024gemini1_5}. 
These advancements highlight the diverse and expanding landscape of MLLMs in the field, which has remarkably impacted the landscape of AGI.

\subsection{Visual Projector in MLLMs}
Visual projector plays a fundamental role to bridge the vision and language model, which aligns visual signals from a visual encoder with the LLM space. Current approaches can be mainly divided into two categories. One is linear projection~\cite{liu2023llava, llava1.5} through MLP. MLP projection can preserve all visual contexts through a one-to-one transformation, which retains the detailed information with redundant tokens~\cite{bolya2022token,2024llava-prumerge}. A critical concern with this method is the substantial increment in the number of visual tokens, especially in processing high-resolution images or videos. 
To tackle this issue, another research line focuses on the reduction of visual tokens to improve the efficiency of MLLMs. Resampler~\cite{bai2023qwen-vl} or  Q-Former~\cite{BLIP2} employs learnable queries to explicitly controls the number of visual tokens and  force extracting the most relevant visual cues from visual features by cross-attention layers. Building on Resampler, Yu \textit{et al.}~\cite{yu2024texthawk} propose a Query Proposal Network (QPN) to generate the initial query and perform the multi-level cross attention.
Some recent works, \textit{e.g.} Abstractor~\cite{cha2023honeybee} and LDP~\cite{mobilevlmv1, mobilevlmv2} adopt convolution layers to encourage local interaction of visual features and generate the compressed tokens. However, these methods inevitably omit fine detailed information, thereby compromising visual reasoning abilities of MLLMs.  Additionally, some works directly transfer visual features from the length dimension to channel dimension by a simple pixel shuffle~\cite{chen2024howfar} or nearby concatenation~\cite{dong2024internlm} operation to  reduce the number of visual tokens. Although all information are retained, it may destroy  intrinsic characteristics of the visual feature itself. Recent research~\cite{mckinzie2024mm1} has undertaken an empirical study on commonly-used projectors, concluding that their types have negligible  effect.  In contrast to these findings, this paper introduces a novel and effective visual projector dubbed TokenPacker.

\subsection{High-Resolution Understanding with MLLMs}
Most of MLLMs commonly utilize CLIP-ViT~\cite{CLIP} as the visual encoder to capture visual information. However, the vision encoder is constrained by low-resolution input, such as 224$\times$224 or 336$\times$ 336, which impedes the ability of MLLMs to effectively manage tasks that require finer details, like dense OCR, crowd counting and visual grounding of small objects. To overcome this limitation, a group of methods~\cite{wei2023vary,hong2023cogagent,yuan2023osprey,li2024miniGemini,lu2024deepseek} directly employ the visual encoder, like SAM encoder~\cite{kirillov2023segment} or ConvNeXt~\cite{liu2022convnet}, that efficiently supports high-resolution input to capture the finer visual cues.  Different from these methods, the patch-cropping strategies are introduced to split a high-resolution image into multiple image patches. The image patches are then processed separately to obtain the visual embeddings of the entire high-resolution image. Some works~\cite{lin2023sphinx, li2023monkey} first resize input image into an accessible size, and adopts the sliding windows to segment images into the uniform patches (\textit{e.g.} 224$\times$224). While these methods change the raw resolution into a fixed square size, this may result in blurring or distortion of visual content. To alleviate this problem, several studies~\cite{ye2023ureader,llavanext,dong2024xc24khd,chen2024howfar} leverage a similar aspect ratio with input image to resize, instead of adhering to a fixed square ratio. 

\begin{figure}[t]
\begin{center}
\includegraphics[width=1.0\linewidth]{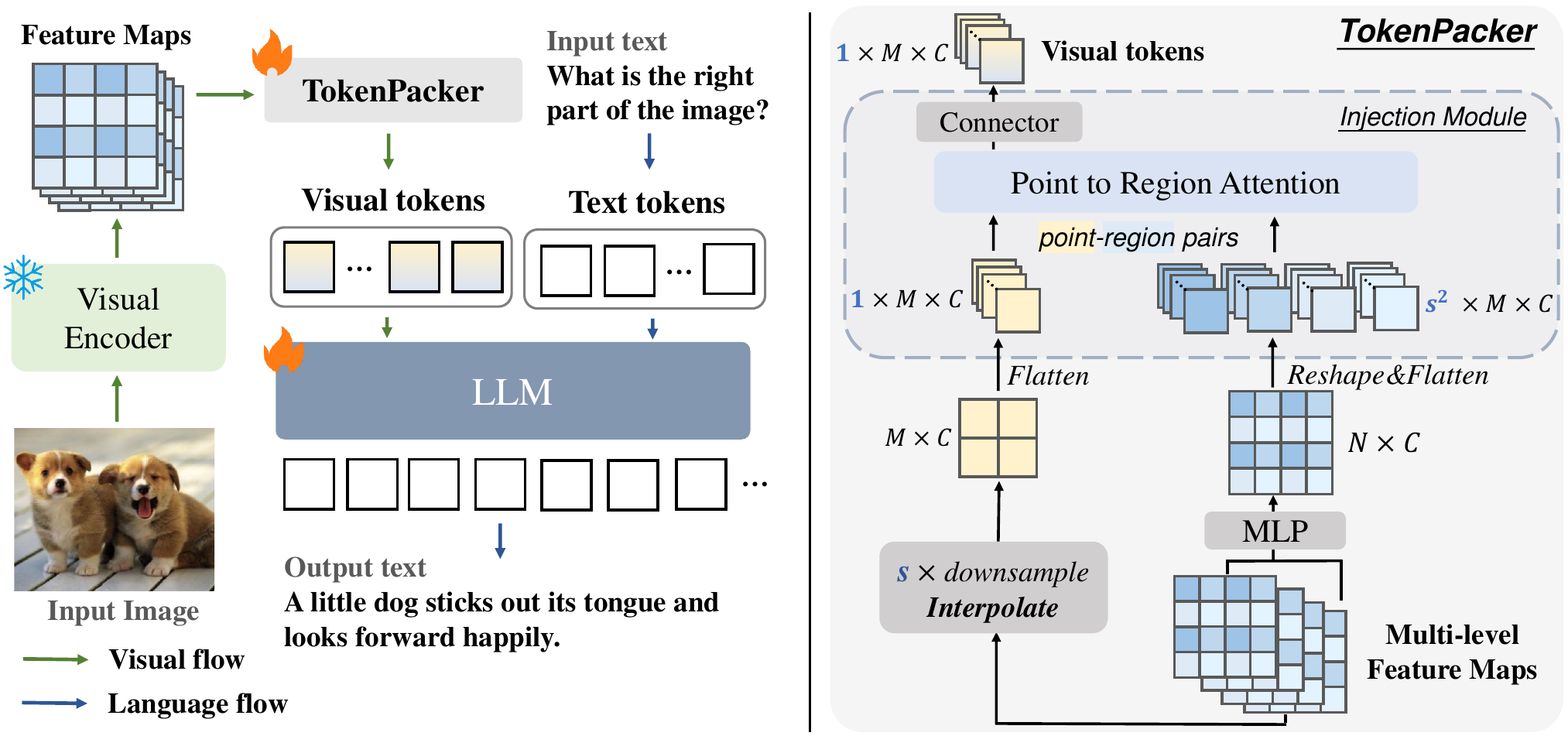} \end{center}
 \vspace{-5pt}
    \caption{(Left) Overview of a standard MLLM framework with our TokenPacker as the visual projector. (Right) The architecture of TokenPacker. TokenPacker initially interpolates visual features as a low-resolution point query. Subsequently, high-resolution and multi-level region cues are treated as reference keys and values to inject their finer information to update coarse query via point to region attention in a local context.  TokenPacker can generate the compact visual tokens in small quantities, yet encapsulate rich details efficiently.}
 \label{fig:framework}
\vspace{-10pt}
 \end{figure}

\section{Method}

In this section, we first revisit the overall framework of a standard MLLM that generates  instruction-following response for the given multimodal inputs (Section~\ref{sec3.1}). Then, we introduce our effective visual projector named TokenPacker, specially designed for bridging visual encoder and LLM, to generate the condensed visual token representations for the subsequent LLM processing (Section~\ref{sec3.2}).  Finally, we present an dynamic image slicing scheme that supports input images in any aspect ratios with a minimal padding content. By integrating TokenPaker, our approach can achieve fine-grained high-resolution image understanding with significant computation efficiency (Section~\ref{sec3.3}).

\subsection{Revisiting Multimodal Large Language Models (MLLMs)}\label{sec3.1}
The aim of MLLMs is to develop a sophisticated model  capable of generating producing responses that adhere to given instructions upon multimodal inputs, including visual and textual data.  MLLMs are typically composed of three pivotal components:  1) Visual Encoder $\mathbf{F}_{I}$: it converts an  input image ${I}_{img}\in\mathbb{R}^{H {\times}W{\times}3}$ into a group of distinctive visual embeddings $ {I}_v\in\mathbb{R}^{N{\times}C}$. It always leverages the widely-used CLIP-ViT-L/14 as its backbone with a patch size $P$ of 14, and $N=HW/{P^2}$ denotes the number of visual embeddings. 2) Visual Projector $\mathbf{\Gamma}_{I\rightarrow T}$: this component translates visual embedding ${I}_v$ into the visual token $\mathbf{T}_v$  in the textual embedding space $T$ with an appropriate dimension for the subsequent language model. 3) LLM  ${\mathbf{\Phi}_{({\mathbf{T}_v},{\mathbf{T}_t})}}$: it takes in both visual token $\mathbf{T}_v$ and textual token $\mathbf{T}_t$, and produces a coherent response auto-regressively. For a sequence of response with length $L$, the probability of generating contextually target answers $\mathbf{Y}=\{y_i\}^L_{i=1}$ can be calculated by:
\begin{equation}
p(\mathbf{Y}|{\mathbf{T}_{v}},{\mathbf{T}_{t}})=\prod_{i=1}^{L}p(y_{i}|{\mathbf{T}_{v}},{\mathbf{T}_{t, {{<i}}}}, {\mathbf{Y}_{{{<i}}}} ).
\end{equation}

In this typical MLLM framework, the computational and memory demands are predominantly dictated by the LLM ${\mathbf{\Phi}_{({\mathbf{T}_v},{\mathbf{T}_t})}}$ with large amount of parameters. It should be emphasized that the computational expenses of LLM  ${\mathbf{\Phi}_{({\mathbf{T}_v},{\mathbf{T}_t})}}$ generally exhibit a quadratic increase relative to the quantity of its input tokens. This highlights the significant impact that the quantity of input tokens has on the overall efficiency of the framework. The visual projector takes the $N$ visual embeddings ${I}_v$ and converted them to $M$ visual tokens $\mathbf{T}_v$. Therefore, reducing the number of visual tokens is a pivotal approach to bolster the efficiency of LLM, \textit{i.e.} $M<N$.

\subsection{TokenPacker: an Efficient Visual Projector} \label{sec3.2}
Visual projector plays a vital role in translating the $N$ visual embeddings ${I}_v$ into $M$ visual tokens $\mathbf{T}_v$ before feeding into LLM.  
As shown in Figure~\ref{fig:framework}, we introduce an effective visual projector, namely TokenPacker, which connects the vision encoder and language model using as small number of tokens as possible. The architecture of our TokenPacker is crafted with a coarse-to-fine framework.

Specifically, we initially downsample the visual features ${I}_v\in\mathbb{R}^{N{\times}C}$ 
before the last Transformer layer of CLIP-based vision encoder
via bilinear interpolation with a scaling factor $s$ as the low-resolution visual embeddings ${I}'_v\in\mathbb{R}^{M{\times}C}$, where $M=N/{s}^2$. Therefore, the number of visual token $M$ can be controlled by the down-sampling ratio $s$. The low-resolution ${I}'_v$ can be regarded as the coarse representations of original high-resolution visual features, where each pixel of low-resolution ${I}'_v$ corresponds to a specific ($s{\times}s$) sub-region of the high-resolution ${I}_v$. 
Subsequently, we construct the point-region pairs, \textit{i.e.} each pixel in ${I}'_v\in\mathbb{R}^{1{\times}M{\times}C}$
to sub-region in  ${I}_v\in\mathbb{R}^{{{s}^2}{\times}M{\times}C}$, and aim to infuse the detailed information of high-resolution sub-region into each pixel with coarse representation. To accomplish this process, we devise an injection module that effectively performs region-to-point information injection to enhance and update the low-resolution representations.

In particular, we take the low-resolution ${I}'_v\in\mathbb{R}^{1{\times}M{\times}C}$ as point-based  queries, and ${I}_v\in\mathbb{R}^{{{s}^2}{\times}M{\times}C}$ as region-based candidate keys and values for reference. The region-to-point information injection is conducted by a point to region cross-attention operation following a MLP layer to make the low-resolution queries fully absorb the fine-grained keys and values and update to be a compact and enhanced visual tokens $\mathbf{T}_v$. Furthermore, we leverage  multi-level visual features as the more enriched reference keys and values. As evidence in prior work~\cite{jiang2023clip}, different layers of CLIP encoder display varying biases towards different patterns.  The shallow layer features contain detailed low-level information, while deep layer features are superior at semantic understanding. The multi-level region-to-point injection process encourages to infuse the plentiful high-resolution information from multiple layers into low-resolution queries, being sufficient to serve as visual tokens. Therefore, our approach is capable of producing superior visual tokens while simultaneously reducing the total number of visual tokens to $1/{s^2}$ of the visual embeddings. 

\begin{figure}[t]
\begin{center}
\includegraphics[width=1.0\linewidth]{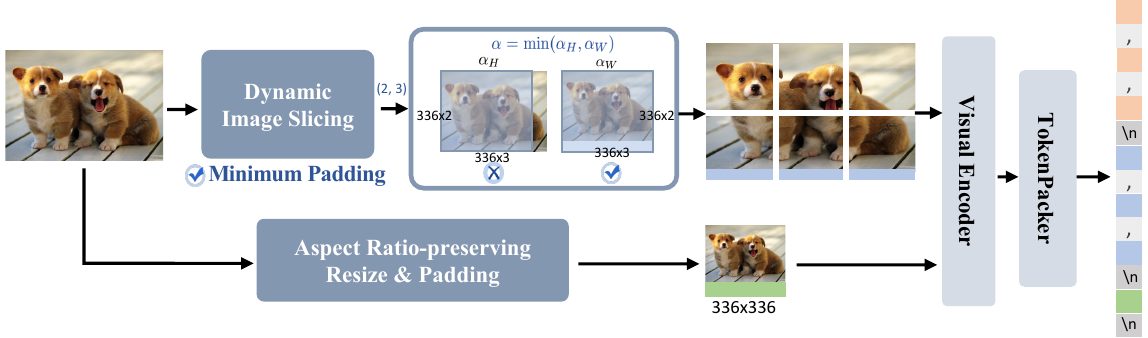} \end{center}
 \vspace{-5pt}
    \caption{The pipeline for efficient high-resolution image understanding with our TokenPacker.}
 \label{fig:partition}
\vspace{-15pt}
 \end{figure}
 
\subsection{High-Resolution Image Understanding with TokenPacker} \label{sec3.3}

To support efficient high-resolution image understanding, we further develop an effective image cropping method with our TokenPacker.  Inspired by previous work~\cite{ye2023ureader}, we focus on an aspect ratio-preserving slicing scheme to avoid the deformation and distortion of visual content that results from resizing operations. Different from the prior approaches~\cite{ye2023ureader,llavanext,dong2024xc24khd,chen2024howfar}, we suggest a dynamic image slicing scheme to preserve any aspect-ratio with the minimum padding as possible to ensure the splitted grid is maximally filled with original image content.

Initially, we specify a set of grids $\mathbb{G}=\{(n_H, n_W)\mid n_H\times n_W \leq N_g, n_H\in \mathbb{N}, n_W\in \mathbb{N}\}$ with various partition configurations for input images.  Here, $n_H$ and $n_W$ are the number of rows and columns of the grids, and $N_g$ denotes the maximum  allowable  number of grids. 
To obtain an optimal grid configuration for a given image ${I}_{img}\in\mathbb{R}^{H {\times}W{\times}3}$, we mainly consider three critical factors: 1) preserving the image's original aspect ratio to avoid distortion; 2) minimizing  the padding proportion so that most of the grids are occupied by original image content; and 3) among the options meeting the first two points, choosing the one whose resolution aligns most closely with the image.

To fulfill the above conditions, we define the padding score $S_p$ and overlap score $S_o$ as following: 
\vspace{-2mm}
\begin{equation}
{S_p(H,W,r,n_H,n_W)} = \frac{{H\times W\times \alpha^2}}{{{n_H} \times {n_W} \times {r^2}}}, 
\end{equation}
\begin{equation}
{S_o(H,W,r,n_H,n_W)} = IoU((H,W),({n_H} \times r,{n_W} \times r)),
\end{equation}
where $\alpha$ is the minimum of two ratios, \textit{i.e.}, $\alpha  = \min ({\alpha_H},{\alpha_W})$. Specifically, ${\alpha_H} = \frac{{{n_H} \times r}}{H}$ and ${\alpha_W} = \frac{{{n_W} \times r}}{W}$. 
$r$ denotes the size of each grid, we set 336 using CLIP-ViT-L/14 as vision encoder. Accordingly, the grids suitable for an image can be identified as follows,
\begin{equation}\label{eq4}
{n_H}^*, {n_W}^* = \mathop{\arg \max}\limits_{(n_H, n_W)\in \mathbb{G}} {S_p(H,W,r,n_H,n_W)} + \beta {S_o(H,W,r,n_H,n_W)}.
\end{equation}
\vspace{-4mm}

As shown in Fig~\ref{fig:partition}, we can obtain image grids of varying sizes with proper configuration for slicing. Then, we resize  raw image by the ratio $\alpha$ and pad the remaining part with zero. To preserve the integrity of original image, we also integrate resized original image with aspect ratio preserving to provide a macroscopic overview as in previous works~\cite{ye2023ureader,llava1.5}. Following the feature extraction of these image patches, our TokenPacker generates the compact visual tokens for each splitted grid and merge to a sequence of visual tokens according to its original
arrangement. Besides, we introduce the comma (\texttt{`,'}) among each grid, and present a newline (\texttt{`\textbackslash {n}'}) token at the end of each row of the image grids to clarify the 2D structure information of image and avoid the ambiguity in the LLM.

\section{Experiments}

In this section, we first introduce the details of our experimental setup. Then, we benchmark our approach against leading methods across various multimodal testbeds. At the end of this section, the ablation analysis and qualitative results are presented.

\subsection{Implementation Details}
In this work, we instantiate our approach  on the top of LLaVA-1.5~\cite{llava1.5}. Specifically, we employed CLIP-ViT-L/14-336px~\cite{CLIP} as visual encoder with the default resolution of 336$\times$336, and adopted Vicuna-7/13B model~\cite{zheng2023vicuna} as the LLM.  We perform a two-stage training paradigm, consisting of a pre-training phase and an instruction-tuning phase.  
To ensure efficiency in training, we maintain  vision encoders fixed across both stages, while focusing on optimizing our proposed TokenPacker. 
Concurrently, the optimization of the LLM is exclusively conducted during the instruction-tuning phase. We adjust the down-sampling ratio $s \in \{2, 3, 4\}$ in TokenPacker to control the quantity of generated visual tokens. In the dynamic slicing scheme for high-resolution image, we set $N_g=9$ or $N_g=16$  for model training  and evaluation to support a range of resolutions, such as $1344{\times}1344$, $5376{\times}336$, etc. In Eq.~\ref{eq4}, we assign $\beta=0.1$ by default. As in ~\cite{llava1.5}, we train all models for one epoch by leveraging the AdamW optimizer with a Cosine learning rate schedule. We set the initial learning rates for pre-training phase and instruction tuning phase at 1$e^{-3}$ and 2$e^{-5}$, respectively. The models are trained on 8 $\times$ NVIDIA A100 GPUs.

\subsection{Datasets and Benchmarks}
To make a fair comparison, we first conduct the experiments on CC-558K dataset~\cite{liu2023llava} for training our TokenPacker in order to perform the modality alignment at the first stage. The 665K mixture dataset~\cite{liu2023llava} is employed for instruction-tuning at the second stage, following LLaVA-1.5~\cite{llava1.5}.  To achieve the competitive performance, we then adopt more high-quality training samples as organized in Mini-Gemini~\cite{li2024miniGemini}, around 1.2M for the first stage and 1.5M for the second stage. Furthermore, we conduct an extensive  evaluation across a series of widely-used benchmarks to assess multimodal understanding and reasoning capability of our proposed model. The benchmarks employed in our study  consist of: 1) General visual question answering benchmarks, like VQA$^\text{v2}$~\cite{goyal2017making},  GQA~\cite{gqa}, VizWiz~\cite{VizWiz}; 2) OCR-related benchmarks, like VQA$^\text{T}$ (TextVQA)~\cite{textvqa}, OCRBench (OCRB)~\cite{liu2023ocrbench} and DocmentVQA(DocVQA)~\cite{mathew2021docvqa}; 3) Hallucination benchmark like POPE~\cite{pope}; 4) Comprehensive benchmarks like MMB (MMBench)~\cite{mmbench},  MM-Vet~\cite{mmvet} and MMMU~\cite{yue2023mmmu}.

\newcommand{\treshl}[2]{
{#1} \fontsize{7.5pt}{1em}\selectfont\color{mygreen}{$\!\uparrow\!$ \textbf{#2}}
}
\newcommand{\tdownreshl}[2]{
{#1} \fontsize{7.5pt}{1em}\selectfont\color{midblue}{$\!\downarrow\!$ \textbf{#2}}
}

\newcommand{\reshl}[2]{
\textbf{#1} \fontsize{7.5pt}{1em}\selectfont\color{mygreen}{$\!\uparrow\!$ \textbf{#2}}
}
\newcommand{\downreshl}[2]{
\textbf{#1} \fontsize{7.5pt}{1em}\selectfont\color{midblue}{$\!\downarrow\!$ \textbf{#2}}
}

\begin{table*}[t!]
\setlength{\tabcolsep}{3.2pt}
\centering
\caption{Comparison with leading methods on zero-shot benchmarks. Our TokenPacker compresses the visual tokens from 576 to 144 (1/4), 64 (1/9) or 36 (1/16) while still delivering competitive performance in comparison to LLaVA-1.5. The results of our method are highlighted with \colorrect{mygray}. }
\vspace{-1.5mm}
\renewcommand\arraystretch{1.16}
\scalebox{0.74}{
\begin{tabular}{l l P{7mm}  P{8mm}  P{8mm}  P{8mm} | p{11mm}p{12.8mm}p{11.0mm}p{11mm}p{11mm}p{11mm}p{6mm}}
\hline\thickhline
Method & LLM & Res. & \#Token & PT & IT & {\bf MMB} & {\bf MM-Vet}  & {\bf VizWiz} & {\bf VQA$^\text{v2}$} & {\bf GQA}  & {\bf POPE}  & {\bf Avg.}  \\
\hline
\hline
MobileVLM V2~\cite{mobilevlmv2} & MLLaMA-2.7B & 336  & 144 &  1.2M & 3.6M&  57.7 & -- & -- & -- & 61.1 & 84.7  & --    \\
Shikra~\cite{chen2023shikra} & Vicuna-13B & 224 & 256 & 600k & 5.5M  &  58.8 & -- & --  & 77.4 & -- & -- & -- \\
IDEFICS-80B~\cite{idefics} & LLaMA-65B & 224 & 256 & 353M & 1M & 54.5 & -- & -- & 60.0 & -- & --  & --  \\
Qwen-VL~\cite{bai2023qwen-vl} & Qwen-7B & 448 & 256  &  1.4B & 50M  & 38.2  & -- & 35.2 & 78.8 & 59.3 & --   & --  \\
Qwen-VL-Chat~\cite{bai2023qwen-vl} & Qwen-7B  & 448  & 256 & 1.4B & 50M &  60.6 & --  & 38.9  & 78.2 &57.5  & -- & --  \\
\hline
LLaVA-1.5~\cite{llava1.5} & Vicuna-7B & 336 & 576 & 558K & 665K  & 64.3 & 31.1  & 50.0 & \textbf{78.5} & \textbf{62.0}  & 85.9 & 62.0   \\
\rowcolor{mygray}
LLaVA-TokenPacker &  Vicuna-7B & 336 &  144  & 558K & 665K & \reshl{65.1}{0.8}  & \reshl{33.0}{1.9}  & \reshl{52.0}{2.0} & \tdownreshl{77.9}{0.6} & \tdownreshl{61.9}{0.1}  & \reshl{87.0}{1.1}  & \textbf{62.8} \\
\hline
LLaVA-1.5~\cite{llava1.5} & Vicuna-13B & 336 &  576 &  558K & 665K & 67.7 & \textbf{36.1} & 53.6  & \textbf{80.0} &  \textbf{63.3} & 85.9  & 64.4  \\
\rowcolor{mygray} 
LLaVA-TokenPacker & Vicuna-13B & 336  &144 &  558K & 665K & \reshl{68.0}{0.3} & \tdownreshl{34.5}{1.6} & \reshl{55.6}{2.0}  & \tdownreshl{78.9}{0.1} & \tdownreshl{62.5}{0.8} & \reshl{87.4}{1.5} & \textbf{64.5}
\\
\hline
\multicolumn{13}{c}{\em  Fewer Tokens Setting}\\
\hline
InstructBLIP~\cite{instructblip} & Vicuna-7B & 224  & 64 & 129M & 1.2M & 36.0 & 26.2 & -- & --& -- & --   & --  \\
InstructBLIP~\cite{instructblip} & Vicuna-13B & 224  & 64 &  129M & 1.2M  &  -- & 25.6  & 33.4 & -- & 49.5 & 78.9  & --    \\
\rowcolor{mygray1}
LLaVA-TokenPacker  & Vicuna-7B & 336 & 64 & 558K & 665K &  64.1 & 31.7 & 50.7 & 77.2 & 61.1 & 86.3 & 61.9 \\
\rowcolor{mygray1}
LLaVA-TokenPacker  & Vicuna-13B & 336 & 64 & 558K & 665K &  \textbf{66.2} & \textbf{34.2} & \textbf{52.9} & \textbf{78.1} & \textbf{62.0} & \textbf{87.3} & \textbf{63.5}\\
\hline
LLaVA-PruMerge~\cite{2024llava-prumerge} & Vicuna-7B & 336 & \textasciitilde32 & 558K & 665K & 60.9 & -- & -- & 72.0 & -- & 86.3 & --\\
LLaVA-PruMerge~\cite{2024llava-prumerge} & Vicuna-13B & 336 & \textasciitilde32 & 558K & 665K & 62.3 & -- & -- & 72.8 & -- & 86.2 & -- \\
\rowcolor{mygray1}
LLaVA-TokenPacker  & Vicuna-7B & 336 & 36 & 558K & 665K &  62.8 & 29.6 & 50.2 & 75.0 & 59.6 & 86.2 & 60.6\\
\rowcolor{mygray1}
LLaVA-TokenPacker  & Vicuna-13B & 336 & 36 & 558K & 665K &  \textbf{66.2} & \textbf{34.1} & \textbf{53.9} & \textbf{76.3} & \textbf{60.7} & \textbf{86.5} & \textbf{63.0} \\
\hline\thickhline
\end{tabular}
}
\vspace{-6mm}
\label{tab:main_result_normal_resolution}
\end{table*}

\subsection{Main Results}

\textbf{Normal Resolution.} We first examine the effectiveness of our proposed TokenPacker in normal resolution settings with the data as in LLaVA-1.5~\cite{llava1.5}. We compare our approach with the previous leading methods, including MobileVLM V2~\cite{mobilevlmv2},  Shikra~\cite{chen2023shikra}, IDEFICS~\cite{idefics}, Qwen-VL~\cite{bai2023qwen-vl}, and InstructBLIP~\cite{instructblip}, LLaVA-PruMerge~\cite{2024llava-prumerge} with fewer visual tokens. Six popular benchmarks are adopted including comprehensive MMBench and MM-Vet, and general VQA-related VizWiz, VQA$^\text{v2}$, GQA and hallucination POPE for a thorough performance evaluation. 
As shown in Table~\ref{tab:main_result_normal_resolution}, our approach showcases the superior performance on the MMBench, VizWiz and POPE benchmarks, respectively. When juxtaposed with the baseline LLaVA-1.5 model, our proposed TokenPacker as the visual projector achieves a reduction of visual tokens by 75\% (from 576 to 144), while enhancing performance metrics by +0.8\%/+0.3\% on MMBench, +2.0\% on both measures for VizWiz, and +1.1\%/+1.5\% on POPE with the Vicuna-7B/13B LLMs, respectively. Although a marginal decline in performance on image question answering benchmarks such as VQA$^\text{v2}$ and GQA, our methods still comprehensively brings the average performance gains over LLaVA-1.5~\cite{llava1.5}, +0.8\% and +0.1\%  respectively using Vicuna-7B and Vicuna-13B models with around 5 times TPS (4.9 \textit{vs.} 24.9, see Table~\ref{tab:resampler} for the details). Additionally, our approach exceeds the previous methods, like Qwen-VL-Chat~\cite{bai2023qwen-vl}, InstructBLIP~\cite{instructblip} and MobileVLM V2~\cite{mobilevlmv2} across most of benchmarks, regardless of their access to more substantial training data. 

Furthermore, we compare our method against previous leading approaches with fewer visual tokens. Specially, we set the token number to 64 (11\% of 576) and 36 (6\% of 576), respectively. It can be seen that our method  surpasses these methods across three benchmarks at a large margin. For example, we observe that +3.9\% on MMBench, +3.5\% on VQA$^\text{v2}$ are achieved against recent LLaVA-PruMerge~\cite{2024llava-prumerge} with Vicuna-13B. These results affirm the efficacy of our TokenPacker, underscoring its advantageous impact on enhancing visual token representation and overall performance.

\begin{table}[t]
\setlength{\tabcolsep}{2.1pt}
\centering
\caption{Performance comparisons with high-resolution approaches on nine multimodal benchmarks.  The token number of our method is the average  statistically across all training and test data. $^\dag$, $^\sharp$ and $^\S$ denote $s=2, 3, 4$ in TokenPacker, respectively. The best results are \textbf{bold} and the second-best results are \underline{underlined}. * denotes the results obtained through the officially public protocols and checkpoints.}
 \vspace{0.8mm}
\renewcommand\arraystretch{1.20}
\scalebox{0.66}{
\begin{tabular}{l l l l c|p{11mm}p{10mm}p{13mm} | p{8.5mm}p{11.5mm}p{13mm}|p{12.0mm}p{11mm}|p{10mm}}
\hline\thickhline
Method & LLM & \#Data & Max Res. & \#Token & {\bf VQA$^\text{T}$} & {\bf OCRB} & {\bf DocVQA} & {\bf MMB} & {\bf MMMU} & {\bf MME} & {\bf VQA$^\text{v2}$} 
 & {\bf VizWiz}   & {\bf POPE} \\
\hline
\hline
OtterHD~\cite{+0li2023otterhd} & Fuyu-8B~\cite{fuyu8b} & - & 1024$\times$1024 & -- & -- & -- & --  &58.3  & -- & 1294/-- & -- & -- & 86.0  \\
SPHINX-2k~\cite{lin2023sphinx} & LLaMA-13B & 1.0B & 762$\times$762 & 2890 & 61.2 & --& -- & 65.9 & -- & 1471/-- & 80.7 & 44.9 & 87.2  \\ 
UReader~\cite{ye2023ureader} & LLaMA-13B & 86M  & 896$\times$1120 &  -- & 57.6 & -- & \underline{65.4} & -- & -- & -- & -- & -- & -- \\
Monkey~\cite{li2023monkey} & QWen-7B & 1.0B & 896$\times$1344 &  1792 &  -- & \underline{514} &  -- &  -- & -- & -- & 80.3 & \textbf{61.2} & 67.6  \\

TextHawk~\cite{yu2024texthawk} & InternLM-7B & 115M & 1344$\times$1344 & -- &-- & --& \textbf{76.4} & \textbf{74.6} & -- & 1500/-- & -- & -- & --\\

LLaVA-UHD~\cite{xu2024llava_uhd} & Vicuna-13B & 1.2M & 672$\times$1008  & -- & 67.7 & -- & -- & 68.0 &  -- &  1535/--  & 81.7 &  56.1 & \textbf{89.1}  \\

LLaVA-NeXT~\cite{llavanext} & Vicuna-7B  & 1.3M  & 672$\times$672 & 2880 &  64.9 & -- & -- &  67.4  & 35.8 & 1519/332  &  81.8 &  57.6  & 86.5 \\ 

LLaVA-NeXT~\cite{llavanext} & Vicuna-13B  & 1.3M  & 672$\times$672 & 2880 &  67.1 & -- & -- & \underline{70.0} &  36.2 & 1575/326 &  \textbf{82.8} & 60.5 & 86.2  \\

Mini-Genimi-HD~\cite{li2024miniGemini} & Vicuna-7B &  2.7M  & 1536$\times$1536 & 2880 &  68.4 & 456* &  65.0* &  65.8 & 36.8  & 1546/319 & 80.3* & 54.6* & 86.8*  \\
Mini-Genimi-HD~\cite{li2024miniGemini} & Vicuna-13B & 2.7M & 1536$\times$1536 & 2880 & \underline{70.2} & 501* & \underline{70.0}* &  68.6 &  37.3 &  1575/326 & 81.5* & 57.2* & 87.0* \\
\hline
\rowcolor{mygray}
LLaVA-TokenPacker-HD & Vicuna-7B  &  2.7M  & 1088$\times$1088 & \textasciitilde 954$^\dag$  & 68.0 &  452 & 60.2 & 67.4  & 35.4 & 1489/338  & 81.2  & 54.7 & \underline{88.2}   \\
\rowcolor{mygray}
LLaVA-TokenPacker-HD & Vicuna-13B  &  2.7M   &  1088$\times$1088 & \textasciitilde  954$^\dag$ & 69.3 & 498 & 63.0 & 69.5 & \textbf{38.8} & \textbf{1595}/\textbf{356}  &\underline{82.0}  & 59.2 & 88.1  \\ 
\rowcolor{mygray}
LLaVA-TokenPacker-HD & Vicuna-13B  &  2.7M   &  1344$\times$1344 & \textasciitilde 1393$^\dag$ &  \textbf{70.6} & \textbf{521} &  \underline{70.0} &  68.7 & 37.4 & 1574/350 &  81.7 &  57.0 & 88.0 \\

\rowcolor{mygray}
LLaVA-TokenPacker-HD & Vicuna-13B  &  2.7M   & 1344$\times$1344 & \textasciitilde 619$^\sharp$ & 68.8 & 470 & 63.0 & 69.9 & \underline{38.2} & \underline{1577}/\underline{353} & 81.7 & \underline{61.0} & 87.6 \\

\rowcolor{mygray}
LLaVA-TokenPacker-HD & Vicuna-13B  &  2.7M   & 1344$\times$1344 & \textasciitilde 347$^\S$ & 68.4 & 447 & 58.0 & 68.3 & 36.9 & \underline{1577}/332  & 81.2 & 58.1 &  88.0  \\

\hline\thickhline
\end{tabular}
}
\label{tab:main_result_high_resolution}
\vspace{-6mm}
\end{table}

\begin{figure}[t]
\begin{center}
\includegraphics[width=0.99\linewidth]{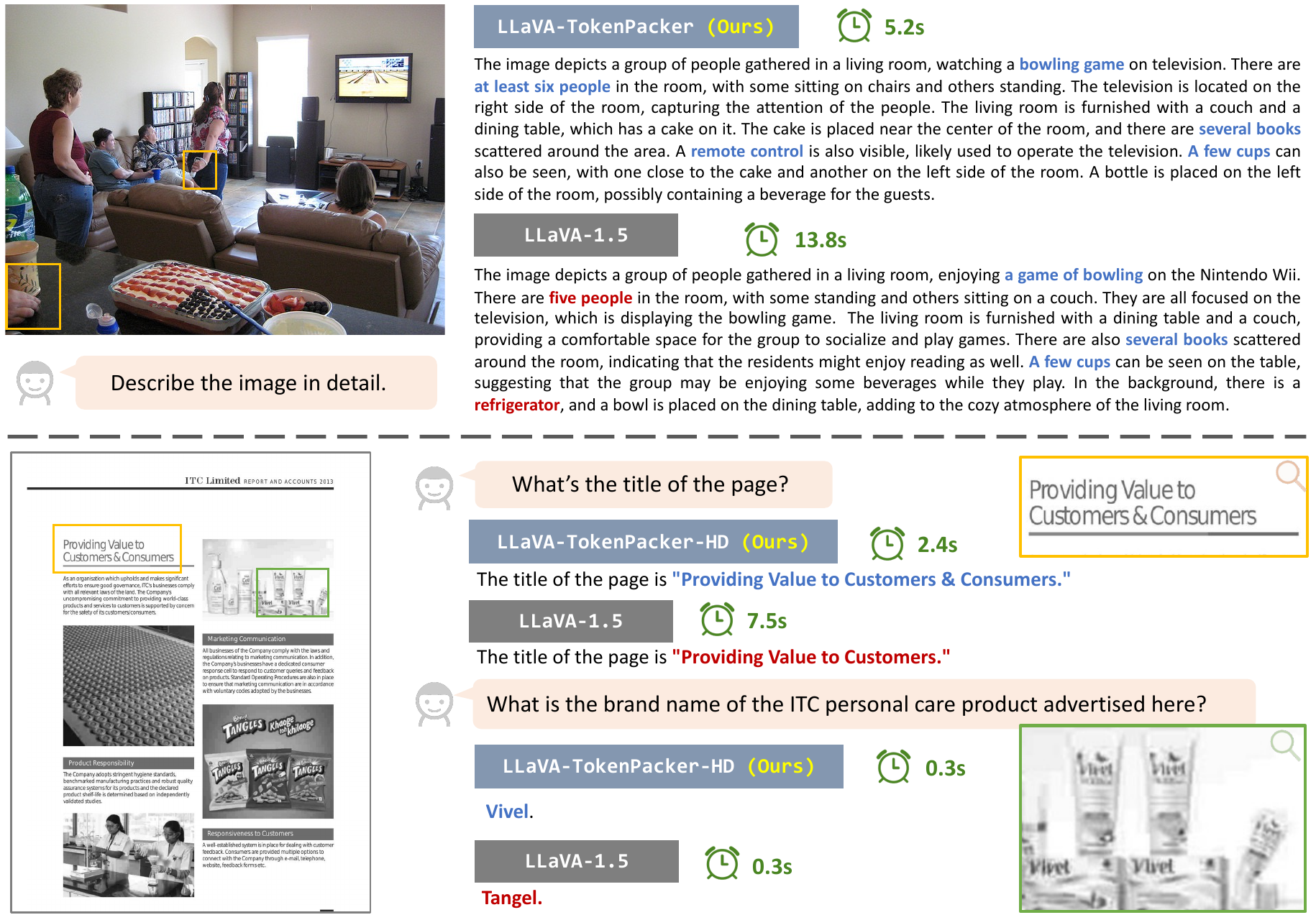} \end{center}
 \vspace{-2.0pt}
    \caption{Qualitative comparisons for representative scenarios. Our approach (144 tokens) achieves to handle the content details correctly and facilitates efficient image understanding. Moreover, our high-resolution method is able to capture the finer elements compared to original LLaVA-1.5.}
 \label{fig:example}
\vspace{-14pt}
 \end{figure}

\textbf{High Resolution.} We apply our methods including TokenPacker and dynamic image slicing scheme into LLaVA-1.5  (dubbed LLaVA-TokenPacker-HD) to perform the high-resolution image understanding. For model training, the 2.7M data  organized in Mini-Gemini~\cite{li2024miniGemini} are employed. We set $N_g=9$ and $N_g=16$ to support the maximum input resolution with 1088$\times$1088 and 1344$\times$1344, respectively. The down-sampling ratio $s$ is set to $2, 3$ or  $4$ to control the quantity of visual tokens derived  from each image patch.  We compare our approach against the existing  high-resolution MLLM methods, such as  OtterHD~\cite{+0li2023otterhd}, SPHINX-2k~\cite{lin2023sphinx}, Monkey~\cite{li2023monkey}, document-oriented UReader~\cite{ye2023ureader} and  TextHawk~\cite{yu2024texthawk}, and the more recent LLaVA-UHD~\cite{xu2024llava_uhd}, LLaVA-NeXT~\cite{llavanext}, Mini-Gemini-HD~\cite{li2024miniGemini}. Table~\ref{tab:main_result_high_resolution} reports the comparison results on nine popular benchmarks, including OCR-related VQA$^\text{T}$, OCRB and DocVQA, and comprehensive MMB, MMMU and MME, and general VQA-related VQA$^\text{v2}$,  VizWiz  and POPE benchmarks. It can be seen that our method with Vicuna-13B as LLM achieves the state-of-the-art OCR-realted performance of 70.6\% on VQA$^\text{T}$ and 521 on OCRBench, when the input resolution is set to 1344$\times$1344 with approximately 1393 visual tokens. 
These promising results can be attributed to the fact that the high-resolution images facilitate MLLM with more visual tokens to precisely recognize intricate fine-grained optical characters or objects. However, for comprehensive benchmarks like MMMU and MME, our approach exhibits the best performance at a lower resolutions with 1088$\times$1088. Besides, even with approximately 619 visual tokens, our method obtains the second-best MMMU, MME and VizWiz scores with 38.2\%, 1577/353, and 61.0\%, respectively. This results demonstrate that MLLM with a reduced number of tokens still deliver robust performance on comprehensive benchmarks and VQA-related tasks.  These results demonstrate the pivotal effectiveness of leveraging native high-resolution imagery in diverse multimodal tasks, and highlight the efficacy of our proposed TokenPacker. Figure~\ref{fig:example} shows the qualitative comparisons across the representative scenarios.  
%, and 70.0\% on DocVQA

\begin{table}[t]
\setlength{\tabcolsep}{4.5pt}
\centering
\caption{Evaluation results on different visual projectors. We adopt token per second (TPS) to evaluate the throughput of LLM during inference, measured by a single NVIDIA A100 GPU.}
\renewcommand\arraystretch{1.18}
\vspace{0.5mm} 
\scalebox{0.78}{
\begin{tabular}{l | P{10mm} P{10mm}|P{13mm}P{15mm}P{14mm}P{14mm}P{13mm}P{13mm}|P{10mm}}
\hline\thickhline
Projector & \#Token & TPS & {\bf MMB} & {\bf MM-Vet} & {\bf VQA$^\text{v2}$} & {\bf GQA}  & {\bf POPE} & {\bf VizWiz} & {\bf Avg.}  \\
\hline
\hline
MLP~\cite{llava1.5} & 576 & 4.9 & 64.3 & 31.1 & 78.5 & 62.0 & 85.9 & 50.0 & 62.0 \\
\hline
Average-Pooling  & 144 & \textbf{27.7} & 64.3 & 26.7 & 76.4 & 60.3 & 86.4 & 51.3 & 60.9  \\
Resampler~\cite{bai2023qwen-vl} & 144 & 24.8 & 63.1 & 29.2 & 75.1 & 58.4 & 84.7 & 51.9 & 60.4\\
C-Abstractor~\cite{cha2023honeybee} & 144 & 24.1 & 63.1 & 29.4 & 74.6 & 59.2 & 84.6 & 49.2 & 60.0 \\
Pixel-Shuffle~\cite{chen2024howfar} & 144 & 25.2 & 64.0 & 29.7 & 76.2 & 60.1 & 85.9 & 48.8 & 60.8\\
LDP-v2~\cite{mobilevlmv2} & 144 & 25.1 & \textbf{66.2} & 28.7 & 77.3 & 61.1 & 86.1 & 47.6 & 61.2\\
\rowcolor{mygray}
Ours & 144 & 24.9 & 65.1 & \textbf{33.0} & \textbf{77.9} & \textbf{61.8} & \textbf{87.0} & \textbf{52.0} & \textbf{62.8} \\
\hline
Average-Pooling & 64 & \textbf{29.2} & 62.4 & 27.1 & 72.6 & 58.8 & 85.4 & 48.0 & 59.1 \\
Resampler~\cite{bai2023qwen-vl} & 64 & 26.6 & 63.4 & 29.2 & 74.1 & 57.7 & 83.4 & \textbf{53.0} & 60.1 \\
C-Abstractor~\cite{cha2023honeybee} & 64 & 26.5 & 62.5 & 29.0 & 74.4 & 59.3 & 85.0 & 45.6 & 59.3 \\
Pixel-Shuffle~\cite{chen2024howfar} & 64 & 27.7 & 63.2 & 28.5 & 74.6 & 59.1 & 85.2 & 47.4 & 59.7\\
LDP-v2~\cite{mobilevlmv2} & 64 & 27.1 & 63.7 & 30.0 & 75.3 & 59.7 & 85.5 & 49.3 & 60.6\\
\rowcolor{mygray}
Ours & 64 & 27.1 & \textbf{64.1} & \textbf{31.7} & \textbf{77.2} & \textbf{61.1} & \textbf{86.3} & 50.7 & \textbf{61.9}  \\
\hline\thickhline
\end{tabular}
}
\label{tab:resampler}
\vspace{-2mm}
\end{table}

\subsection{Ablation Results}
We further dive into in-depth ablation studies to analyze the effectiveness on each component of our approach. All ablation experiments are conducted by employing Vicuna-7B as LLM and the data as in LLaVA-1.5~\cite{llava1.5} for model training.

\textbf{Various Visual Projectors.} We first compare our proposed TokenPacker against various previous visual projectors, including direct Average-Pooling, Resampler~\cite{bai2023qwen-vl}, C-Abstractor~\cite{cha2023honeybee}, Pixel-Shuffle~\cite{chen2024howfar} and the recent LDP-v2~\cite{mobilevlmv2} on the top of LLaVA-1.5. For the Average-Pooling approach, we directly interpolate the feature map from visual encoder using average pooling, and then use the MLP to generate visual tokens. 
We replace the original MLP in LLaVA-1.5 with various projectors and keep the same settings to facilitate a fair comparison. 
To reflect the inference speed, we adopt the token per second (TPS) metric to evaluate the throughput of LLM. 
Table~\ref{tab:resampler} reports the comparison results. In comparison to the MLP projector, all other visual projectors effectively reduce the  number of visual tokens with the significant improvement on inference speed (around 5 \textit{vs.} 25 TPS). Average-Pooling achieves the best inference speed with fewer parameters.
Our visual projector achieves +0.8\% average performance gain with 144 tokens against the MLP with 576 tokens. 
Especially on the MM-Vet~\cite{mmvet}, POPE~\cite{pope} and VizWiz~\cite{VizWiz} benchmarks, our method outperforms MLP-based method by +1.9\%, +1.1\% and +2.0\%, respectively. 
Comparing to other methods, our approach surpasses the previous best method LDP-v2~\cite{mobilevlmv2} by +1.6\%.  In the scenario with 64 tokens, our approach attains a 61.9\% average performance, on par with MLP-based method (61.9\% \textit{vs.} 62.0\%) and surpasses LDP-v2~\cite{mobilevlmv2} by +1.3\%.  These results demonstrate the effectiveness of TokenPacker compared to previous approaches. 

\setlength\intextsep{0pt}
\begin{wraptable}{r}{0.53\linewidth}
	\centering
	\setlength{\abovecaptionskip}{0cm}
    \captionsetup{width=.53\textwidth}    
\caption{Experimental results with different image splicing schemes.}
{\hspace{-0.1ex}
\resizebox{0.53\textwidth}{!}{
    \setlength\tabcolsep{2pt}
    \renewcommand\arraystretch{1.18}
\begin{tabular}{l | l| P{10mm}  p{10mm} p{11mm} p{12mm}} 
\hline\thickhline
 Method & Res. &{\bf VQA$^\text{v2}$} & {\bf GQA}  & {\bf VQA$^\text{T}$}  &  {\bf OCRB}  \\
\hline
FixedSplit~\cite{llava1.5} & 672 & 79.5 & \textbf{63.4} & 62.4 & 327  \\
AdaptiveSplit~\cite{ye2023ureader} & Any & 79.6 & 62.8 & 63.4 & 332  \\
\rowcolor{mygray}
Ours & Any & \textbf{79.9} & 63.2 &  \textbf{64.0} &  \textbf{336}   \\
\hline\thickhline
\end{tabular}}}
\label{tab:ablation_slicing}
\vspace{0.8mm}
\end{wraptable}

\textbf{Different Image Slicing Schemes.} We then compare our dynamic image slicing scheme with the existing approaches for high-resolution image. Here we list two typical methods in previous works. The first one is presented in LLaVA-1.5-HD~\cite{llava1.5}. It first resizes the original image into a fixed larger resolution (e.g. 672$\times$672 ), then divides the image into smaller image patches. For the sake of brevity, we refer to this approach as ``FixedSplit''.  The second one is the shape-adaptive cropping scheme presented in UReader~\cite{ye2023ureader}. This module also considers to preserve the resolution of the image and the cropping grid fits the aspect ratio of the input image. Nevertheless, there still exists the resize operation in a small scale without considering the quantity of padding. We denote this method  as ``AdaptiveSplit'' for clarity. To facilitate a fair comparison, we re-implement both methods by adopting our TokenPacker as visual projector. As shown in Table~\ref{tab:ablation_slicing}, our proposed dynamic image slicing scheme outperforms the previous methods on most of  benchmarks.  In particular, as for OCR-related benchmarks such as VQA$^\text{T}$, OCRBench, the ratio-preserving approaches including AdaptiveSplit~\cite{ye2023ureader} and our method, surpass FixedSplit by +1.0\%/+1.6\% and +5/+9, respectively.

\setlength\intextsep{0pt}
\begin{wraptable}{r}{0.52\linewidth}
	\centering
	\setlength{\abovecaptionskip}{0cm}
    \captionsetup{width=.52\textwidth}    
\caption{Component-wise ablation results. }
{\hspace{-0.1ex}
\resizebox{0.52\textwidth}{!}{
    \setlength\tabcolsep{2pt}
    \renewcommand\arraystretch{1.20}
\begin{tabular}{l | P{12mm} | p{13mm} p{13mm} p{13mm}}
\hline\thickhline
 Method & \#Token & {\bf VQA$^\text{v2}$} & {\bf GQA}  & {\bf VQA$^\text{T}$}  \\
\hline
Baseline & 144 & 76.4 & 60.3 & 55.3  \\
\rowcolor{mygray}
\textbf{\textcolor{darkgreen}{+}} {\em Injection} & 144 & \treshl{77.5}{1.1} & \treshl{61.6}{1.3} & \treshl{56.5}{1.2} \\
\textbf{\textcolor{darkblue}{\circled{c}}} {\em Learnable Query} & 144 & \tdownreshl{76.1}{1.4} & \tdownreshl{59.8}{1.8} &  \tdownreshl{55.2}{1.3}\\
\rowcolor{mygray}
\textbf{\textcolor{darkgreen}{+}} {\em Multi-level Feature} & 144 & \treshl{77.9}{0.4} & \treshl{61.9}{0.3} & \treshl{57.2}{0.7} \\
\rowcolor{mygray}
\textbf{\textcolor{darkgreen}{+}} {\em Image Partition}  & -- & \reshl{79.9}{2.0} & \reshl{63.2}{1.3} & \reshl{64.0}{6.8} \\
\textbf{\textcolor{darkblue}{--}} {\em Separator Token } & -- & \tdownreshl{76.6}{3.3} & \tdownreshl{61.1}{2.1} & \tdownreshl{58.3}{5.7}   \\
\hline\thickhline
\end{tabular}
    }
    }
\label{tab:ablation-component}
\vspace{-0.01mm}
\end{wraptable}

\textbf{Component-wise analysis.} 
Table~\ref{tab:ablation-component} reports the component-wise experimental results of our method. Firstly, we directly employ the 2$\times$ downsampling feature map from visual encoder as low-resolution visual embeddings to feed  MLP projector, yielding 144 visual tokens. We set this as the baseline method that achieves 76.4\%, 60.3\% and 55.3\% on VQA$^\text{v2}$, GQA and VQA$^\text{T}$ benchmarks, respectively.  We then add (\textbf{\textcolor{darkgreen}{+}}) our injection module, which infuses high-resolution characteristic into the low-resolution query to be improved. The injection module obtains +1.1\%, +1.3\% and +1.2\% performance gains over the baseline method, respectively.  Subsequently, when we change (\textbf{\textcolor{darkblue}{\circled{c}}}) the query from low-resolution feature map to \textit{a learnable query}, the performance decreases with -1.4\%, -1.8\%, and -1.3\%. 
The results demonstrate that the downsampled low-resolution feature map  provides a foundation for absorbing finer high-resolution features.  
We further employ multi-level visual features as the comprehensive reference keys and values instead of a single-level feature in the injection module. This brings +0.4\%, +0.3\% and +0.7\% improvements, respectively.  Finally, we add our image slicing scheme for high-resolution image understanding, and the model obtains +2.0\%, +1.3\% and +6.8\% improvements. When we remove (\textbf{\textcolor{darkblue}{--}}) the separator token, \textit{i.e.} comma (\texttt{`,'}) and newline (\texttt{`\textbackslash n'}), the results show the performance drops with -3.3\%, -2.1\% and -5.7\%. These results demonstrate the vital role to perverse the 2D image structure information in image slicing scheme. 

% \vspace{-4pt}
\section{Conclusion and Limitation}\label{sec:conclusion}
% \vspace{-4pt}

In this work, we proposed a novel visual projector, namely TokenPacker, for MLLM. Our method followed a coarse-to-fine design, which effectively condensed the enriched high-resolution image features to compact visual tokens.  As an extension, we further presented an effective dynamic image partition scheme to perform efficient high-resolution image understanding. Extensive experiments have been conducted across diverse  benchmarks to verify the effectiveness of our approach. Notably, our TokenPacker can effective reduce 75\%$\sim$89\%  visual tokens in LLaVA-1.5 and maintain comparable or even better performance with significantly higher efficiency. 

\textbf{Limitation.} Our TokenPacker offers commendable performance by compressing visual tokens by up to 89\%, yet it is not entirely without loss. Specifically, when reduced to 32 (6\%) or fewer tokens, a clear decline in performance is evident. We are dedicated to progressing our research to develop more sophisticated visual projectors with very few tokens for efficient visual understanding with MLLM.

{\small
\bibliographystyle{plain}
\bibliography{reference}
}

\clearpage
\appendix
\setcounter{figure}{0}
\setcounter{table}{0}

\renewcommand{\thefigure}{A\arabic{figure}}
\renewcommand{\thetable}{A\arabic{table}}

\section*{Supplemental Material}

In this part, we further provide additional experimental results and more discussions on our approach. The supplementary material is organized as follows:
\begin{itemize}[leftmargin=*]
	\setlength{\itemsep}{0pt}
	\setlength{\parsep}{-0pt}
	\setlength{\parskip}{-0pt}
	\setlength{\leftmargin}{-10pt}
	\vspace{-2pt}
  \item \S\ref{sec:more_results}: additional experimental results; 
  \item \S\ref{sec:broaderimpacts}: broader impacts;
  \item \S\ref{sec:license}: asset license and consent.
\end{itemize}

\section{Additional Experimental Results}\label{sec:more_results}

\subsection{More Ablation study}

\begin{table}[h]
\setlength{\tabcolsep}{5.0pt}
\centering
\caption{Ablation results on various single-level and multi-level features used in TokenPakcer.}
\vspace{1mm}
\renewcommand\arraystretch{1.20}
\scalebox{0.90}{
\begin{tabular}{c P{10mm}  P{14mm} P{14mm} | c  P{10mm}  P{14mm} P{14mm}}
\hline\thickhline
 Single-level & {\bf VQA$^\text{v2}$} & {\bf GQA}  & {\bf VQA$^\text{T}$} & Multi-level & {\bf VQA$^\text{v2}$} & {\bf GQA}  & {\bf VQA$^\text{T}$}  \\
\hline
\textit{23} & 77.5  & 61.6 & 56.5  & \textit{22-23} & 77.5 & 61.4 & 57.1 \\
\textit{22} &   76.3 & 61.3 &  56.7 & \textit{20-21-22-23} & 77.6 & 61.8 & 56.9   \\ 
\textit{20} & 75.8 & 60.8 & 55.7  & \textit{12-16-22-23} & \textbf{77.9} & \textbf{61.9} & \textbf{57.2}   \\ 
\textit{16} &  76.1  & 61.2& 55.5 & \textit{8-12-22-23} & 76.8 & 61.7 &  56.3 \\ 
\hline\thickhline
\end{tabular}
}
\vspace{1mm}
\label{tab:multilayer_1}
\end{table}

In our TokenPacker, the injection module employs multi-level visual features as  high-resolution reference keys and values to enhance the low-resolution query.  To explore its effects and select the suitable combination of multi-level features, we conduct the evaluation experiments. Table~\ref{tab:multilayer_1} reports the comparisons results.  One can see that our method adopting single-level feature from the 23rd layer yields superior average performance compared to other single-level methods. In contrast, features from shallower levels exhibit relatively poorer performance. 
When using the combination of multi-level features from 12th, 16th, 22nd,  and 23rd layers, our method achieves the best performance with 77.9\%, 61.9\% and 57.2\% on  VQA$^\text{v2}$, GQA and VQA$^\text{T}$ benchmarks, respectively. These results highlight the fact that  different layers of CLIP encoder display unique biases towards various patterns ranging from the shallow layer to deep layers. Consequently, an optimal mixture of multi-level features is capable of harnessing a wealth of information, clearly enhancing TokenPacker's effectiveness.

\subsection{Comparisons on Training Times}

\begin{table}[h]
\setlength{\tabcolsep}{5.0pt}
\centering
\caption{Training Times Analysis. Eight NVIDIA A100 GPUs are adopted within a same environment.  The accuracy is averaged across six benchmarks (refer to Table~\ref{tab:main_result_normal_resolution} of main paper).}
\vspace{1mm}
\renewcommand\arraystretch{1.20}
\scalebox{0.90}{
\begin{tabular}{l | c c c | c }
\hline\thickhline
Method & Token &  Pre-training (PT)  &  Instruction-Tuning (IT) & Avg. Acc.   \\
\hline
LLaVA-1.5~\cite{llava1.5}  & 576  & 3.5h  &  10h  & 62.0 \\
LLaVA-TokenPacker  & 144 &   1h & 7.5h  & 62.8  \\ 
LLaVA-TokenPacker  & 64 &  0.7h  & 6.5h& 61.9  \\ 
LLaVA-TokenPacker  & 36 &   0.5h &  6h & 60.6  \\ 
\hline\thickhline
\end{tabular}
}
\label{tab:training-time}
\vspace{1mm}
\end{table}

To illustrate the efficiency of MLLM through our TokenPacker, we further conduct a thorough analysis on  training time. 
We set our TokenPacker with various token quantities and make a comparison against the original LLaVA-1.5~\cite{llava1.5}. Table~\ref{tab:training-time} reports the evaluation results with Vicuna-7B model. As outlined in Table~\ref{tab:training-time},  the evaluation results clearly indicate that our approach consistently requires shorter training times with a fewer number of visual tokens  in comparison to LLaVA-1.5. Specifically, utilizing 36 visual tokens, our approach achieves pre-training and instruction tuning in only 0.5 hours and 6 hours, respectively. These results verify our method's superiority in facilitating efficient MLLM advancements.

\subsection{More Visual Results}

\begin{figure}[t]
\begin{center}
\includegraphics[width=1.0\linewidth]{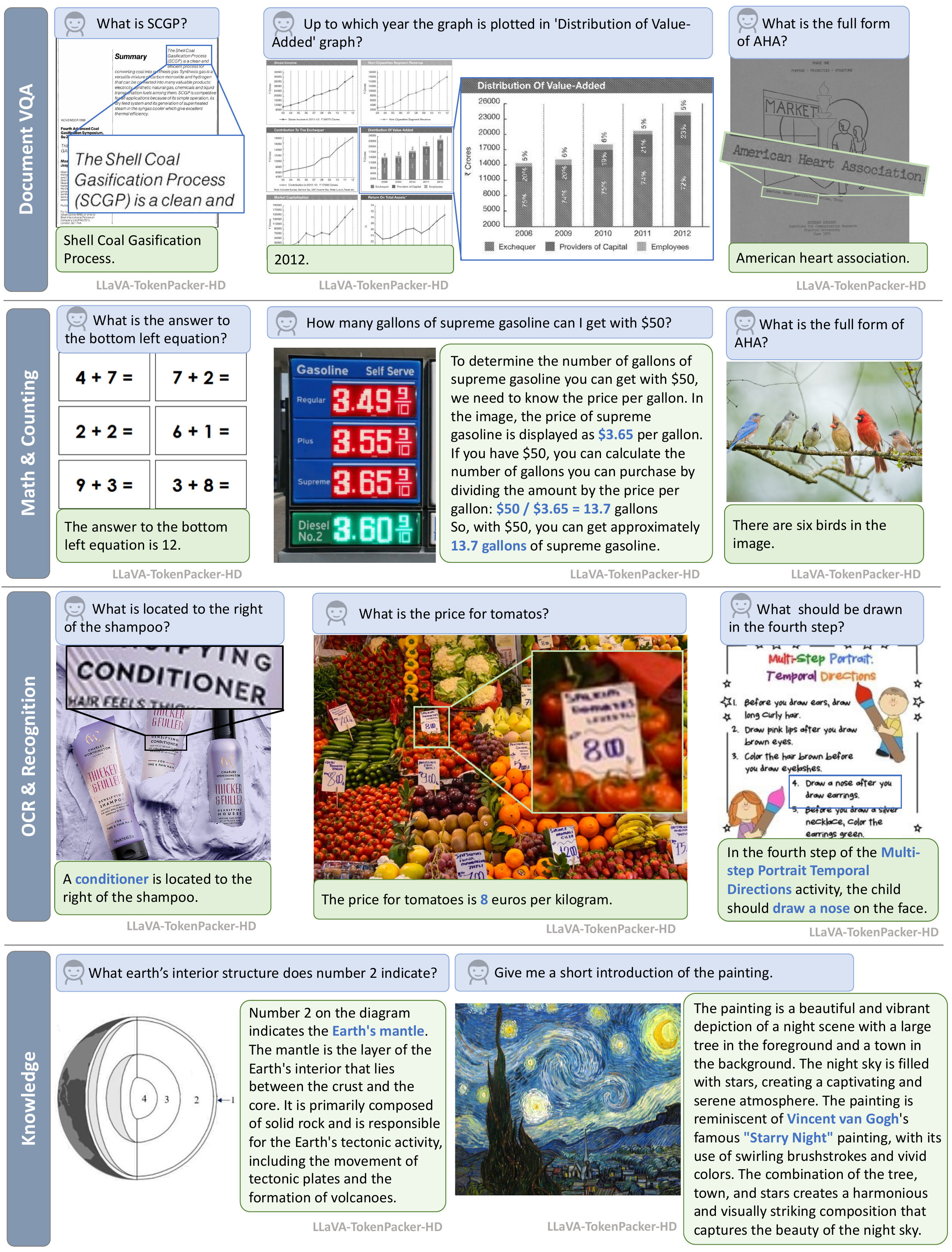} \end{center}
 \vspace{-1pt}
    \caption{Qualitative results across various visual understanding scenarios with our approach.}
 \label{fig:example_vis}
\vspace{-15pt}
 \end{figure}

To verify the effectiveness of our approach for visual comprehension in practical real-world scenarios, we have put it to the test across diverse tasks involving understanding and reasoning, as illustrated in Figure~\ref{fig:example_vis}. Leveraging the capabilities of our TokenPacker combined with a dynamic image slicing scheme for high-resolution image, our method adeptly handles various intricate situations, including document VQA, Math\&Counting, OCR recognition and other tasks that require specialized knowledge.

\section{Broader Impacts}\label{sec:broaderimpacts}

This work presents an effective visual projector for efficient MLLM. We have demonstrated its effectiveness over various multimodal benchmarks. On the positive side, our approach has the potential to benefit the efficient MLLM of real-world image or video understanding, which can clearly reduce the training and inference costs while maintain the competitive performance. On the other side, due to the issue on the robustness of LMMs, some erroneous responses may raise the misinformation or safety   issues of human beings. In order to avoid the potentially negative effects, we suggest to adopt a highly stringent security protocol in case that our approach fails to function properly in real-world multimodal applications.

\section{Asset License and Consent} \label{sec:license}

We utilize 558K image-caption pairs from the LLaVA-filtered CC3M dataset for pre-training and 665K mixture instruction following data for instruction tuning, which are all publicly and freely available for academic research~\cite{llava1.5}.  The 558K pre-training data is the subset of CC3M with BLIP captions~\cite{BLIP2}, which comply with  \href{https://github.com/google-research-datasets/conceptual-captions/blob/master/LICENSE}{license} of  CC-3M (\href{https://ai.google.com/research/ConceptualCaptions/}{\texttt{https://ai.google.com/research/ConceptualCaptions/}}) and  \href{https://github.com/salesforce/BLIP/blob/main/LICENSE.txt}{license} of BLIP(\href{https://github.com/salesforce/BLIP}{\texttt{https://github.com/salesforce/BLIP}}). The 665K mixture insturtion-following data includes publicly available COCO~\cite{lin2014microsoft}, GQA~\cite{gqa}, OCR-VQA~\cite{ocr-vqa},  TextVQA~\cite{textvqa} and Visual Genome~\cite{visualgenome} as the data sources, which is released under the \href{https://creativecommons.org/licenses/by/4.0/legalcode}{CC BY 4.0}. And the GPT-generated multimodal instruction-following data must should abide by the policy (\href{https://openai.com/policies/terms-of-use}{\texttt{https://openai.com/policies/terms-of-use}}) of OpenAI. The 2.7M data organized in Mini-Genimi~\cite{zhu2023minigpt4} is released under the \href{https://creativecommons.org/licenses/by-nc/4.0/deed.en}{CC BY NC 4.0}.
We implement all methods with LLaVA (\href{https://https://github.com/haotian-liu/LLaVA}{\texttt{https://https://github.com/haotian-liu/LLaVA}}) codebase, which are released under the \href{https://apache.org/licenses/LICENSE-2.0}{Apache-2.0 license}.

\end{document}